\documentclass[letterpaper, 10 pt, conference]{ieeeconf}  % Comment this line out if you need a4paper

\IEEEoverridecommandlockouts                              % This command is only needed if 
\usepackage{graphics} % for pdf, bitmapped graphics files
\usepackage{amsmath} % assumes amsmath package installed
\usepackage{amssymb}  % assumes amsmath package installed
\usepackage{booktabs}
\usepackage{multirow}
\usepackage{graphicx}
\usepackage{microtype}
\usepackage[table]{xcolor}
\usepackage{threeparttable}
\usepackage{cite}
\usepackage{caption}
\title{\LARGE \bf
Radar Panoptic Transformer: Semantic-Guided Mask Queries\\
for Dynamic Scene Understanding
}

\author{Mayank Mayank$^{1,2}$, Bharanidhar Duraisamy$^{1,3}$, Rohit Mohan$^{2}$, and Abhinav Valada$^{2}$%
\thanks{$^{1}$Research and Development, Mercedes-Benz AG, Germany.
{\tt\small \{mayank.mayank, bharanidhar.duraisamy\}@mercedes-benz.com}}%
\thanks{$^{2}$Department of Computer Science, University of Freiburg, Germany.
{\tt\small \{mohan, valada\}@cs.uni-freiburg.de}}%
\thanks{$^{3}$Institute for Intelligent Systems, Esslingen University of Applied Sciences, Germany.}%
}

\begin{document}

\maketitle
\thispagestyle{empty}
\pagestyle{empty}

%%%%%%%%%%%%%%%%%%%%%%%%%%%%%%%%%%%%%%%%%%%%%%%%%%%%%%%%%%%%%%%%%%%%%%%%%%%%%%%%
\begin{abstract}

Automotive radar provides long-range sensing and direct Doppler velocity measurements, but its sparse and non-uniformly sampled point clouds pose significant challenges for panoptic segmentation. Dynamic objects often yield only a few detections amid a dominant static background, providing limiting the geometric cues for separating object instances. We propose \emph{Radar Panoptic Transformer (RPT)}, an end-to-end point-based network that incorporates radar-adaptive feature extraction with semantic-guided mask prediction. Its backbone aggregates pose-aligned temporal context through grouped vector attention and learns per-point neighborhood radii conditioned on range, azimuth, and radar cross section (RCS). Point-wise semantic predictions initialize queries at likely foreground detections and guide masked cross-attention through query-point class agreement, producing instance masks without geometric clustering. An auxiliary RCS-consistency loss matches instance-level statistics between predicted and annotated instances. RPT achieves state-of-the-art performance on RadarScenes with $71.6\%$ mIoU and $82.2\%$ PQ. We further evaluate RPT on MAN TruckScenes, where it outperforms the evaluated baselines in a long-range, full-surround radar setting with sensing ranges up to $230\,\mathrm{m}$. RPT achieves real-time end-to-end inference at $31$\,Hz on RadarScenes, exceeding the sensor frame rate.

\end{abstract}

%%%%%%%%%%%%%%%%%%%%%%%%%%%%%%%%%%%%%%%%%%%%%%%%%%%%%%%%%%%%%%%%%%%%%%%%%%%%%%%%

\section{Introduction}
\label{sec:introduction}

Reliable autonomous driving requires a holistic understanding of dynamic scenes, including the semantic structure of the environment and the identification of individual traffic participants. Panoptic segmentation provides this representation by assigning each sensor measurement a semantic class and, for \emph{thing} classes, an instance identifier~\cite{mohan2022perceiving}. Although extensively studied for cameras~\cite{kappeler2024few} and LiDAR point clouds~\cite{sirohi2022efficientlps,marcuzzi2023maskpls}, panoptic segmentation from automotive radar remains less explored. Radar provides distinct sensing capabilities, including direct radial Doppler velocity measurements, long-range operation, and robustness to adverse illumination and weather~\cite{pan2024ratrack,guo2026irgnn}. In particular, Doppler information provides an additional motion cue for distinguishing individual traffic participants, motivating radar-based panoptic segmentation for dynamic scene understanding.

% through long-range operation, robustness to adverse illumination and weather, and direct radial Doppler velocity~\cite{pan2024ratrack,guo2026irgnn}. These properties are particularly relevant for dynamic scene understanding, where doppler information provides an additional cue for separating and associating traffic participants.

\begin{figure}[!t]
    \centering
    \includegraphics[
        width=0.8\columnwidth,
        trim={50 20 20 8},
        clip
    ]{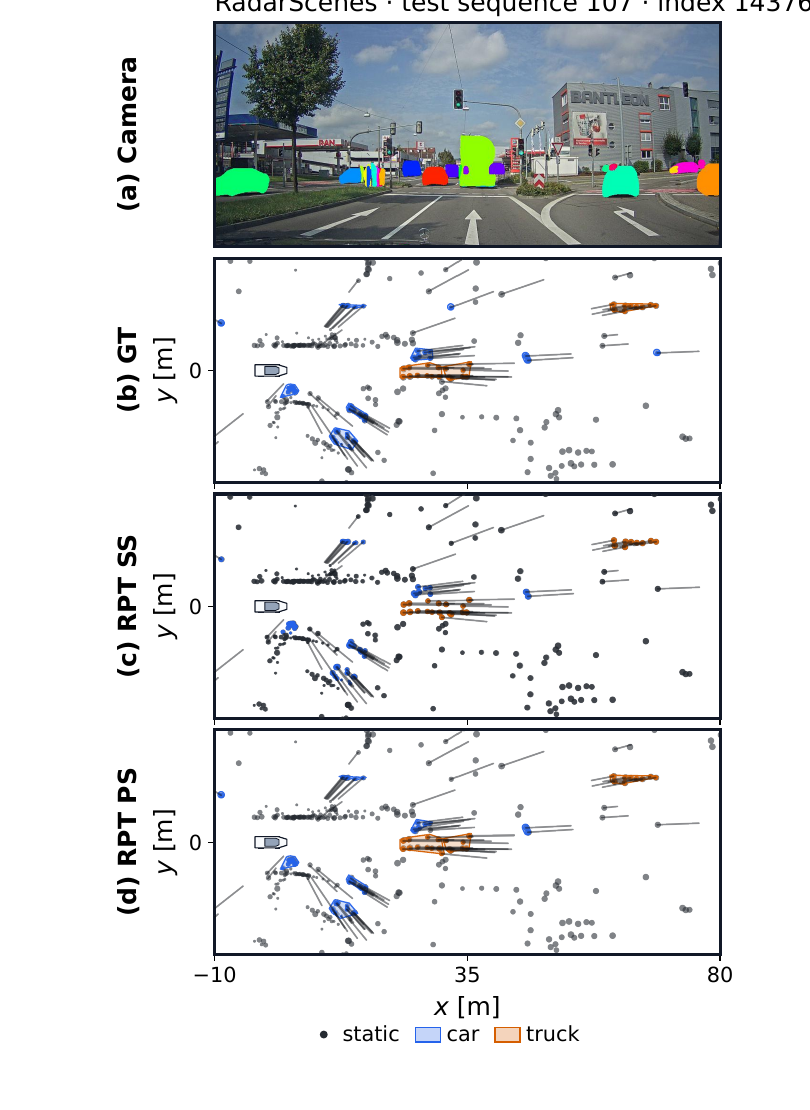}
    \caption{\textbf{Radar Panoptic Transformer (RPT).}
    (a) Camera context, (b) Panoptic ground truth, (c) Semantic prediction, and
    (d) Panoptic prediction on RadarScenes. Semantic predictions guide query-based instance formation for joint point-wise semantic and instance understanding. The scene highlights radar noisy static returns which exhibit spurious high-Doppler measurements and locally resemble dynamic targets. RPT remains robust to these artifacts.}   
    \label{fig:teaser}
    \vspace{-0.8cm}
\end{figure}

Radar perception has predominantly focused on object detection and tracking~\cite{pan2024ratrack,guo2026irgnn,paek2022k}, which yield object-level outputs rather than semantic labels and instance identities for individual radar detections. Achieving such point-level panoptic prediction is challenging as radar measurements are sparse and non-uniformly sampled, with sampling density varying across range, viewing angle, and target reflectivity~\cite{zeller2023grt,zeller2023rvt}. Distant objects may also be represented by only a few detections, while static returns dominate the scene~\cite{schumann2021radarscenes}. Fixed-radius neighborhoods are poorly suited to spatially varying sampling density, while the severe foreground-background imbalance hinders reliable identification of sparsely represented dynamic instances. Radar panoptic segmentation remains a recent research direction~\cite{zeller2025semrafiner}. Existing approaches use refinement pipelines that rely on class-agnostic upstream instance estimates~\cite{zeller2024rit,zeller2025semrafiner}, use fixed-radius neighborhoods, and have largely been evaluated at shorter operating ranges.

We introduce \emph{Radar Panoptic Transformer (RPT)}, an end-to-end point-based architecture for panoptic segmentation of sparse automotive radar point clouds as shown in Fig.~\ref{fig:teaser}. RPT combines KPConv~\cite{thomas2019kpconv} for local geometric encoding with grouped vector attention~\cite{wu2022point} to integrate pose-aligned historical radar context into current-scan features. To account for non-uniform radar sampling, we introduce learned per-point neighborhood radii conditioned on range, azimuth, and RCS. To couple semantic and instance reasoning, we introduce a semantic-guided mask decoder that conditions object-query initialization and masked cross-attention on point-wise semantic predictions, enabling class-aware instance prediction without geometric clustering. We further introduce an RCS-consistency loss that regularizes predicted instances through their instance-level RCS mean and variance. 
%We achieve state-of-the-art performance on RadarScenes~\cite{schumann2021radarscenes}. We further extend radar panoptic evaluation to MAN TruckScenes~\cite{fent2024mantruckscenes}.
We perform extensive evaluations on RadarScenes~\cite{schumann2021radarscenes} and the long-range MAN TruckScenes benchmark~\cite{fent2024mantruckscenes}, where our proposed RPT consistently outperforms existing methods. 

Our contributions are summarized as follows:
\begin{itemize}
    \item We propose \emph{Radar Panoptic Transformer (RPT)}, an end-to-end
    point-based architecture for joint semantic and instance prediction directly from sparse automotive radar measurements.
    \item We introduce a radar-adaptive spatio-temporal encoder that incorporates KPConv~\cite{thomas2019kpconv} with grouped vector attention to integrate pose-aligned historical context into the current-scan representation, together with learned radar-specific neighborhood support that accounts for the non-uniform sampling characteristics of automotive radar.
    \item We propose a semantic-guided mask decoder that uses point-wise
    semantic predictions to initialize and refine object queries, enabling
    class-aware instance formation without explicit geometric clustering. An
    RCS-consistency objective further regularizes the predicted instances using target-level reflectivity statistics.
    \item RPT achieves state-of-the-art panoptic segmentation performance on
    RadarScenes~\cite{schumann2021radarscenes}. We further extend the evaluation to the long-range MAN TruckScenes benchmark~\cite{fent2024mantruckscenes}, where RPT consistently outperforms the evaluated baselines while operating faster than the radar sensor frame rate.
\end{itemize}

% We evaluate RPT on RadarScenes~\cite{schumann2021radarscenes} and MAN TruckScenes~\cite{fent2024mantruckscenes}. RPT achieves state-of-the-art performance on RadarScenes, improving both semantic and instance-level prediction over existing radar approaches. We further extend evaluation to MAN TruckScenes, whose six surround 4D radars, nearly $360^{\circ}$ coverage, and sensing ranges up to $230,\mathrm{m}$ provide a substantially longer-range and full-surround setting. RPT consistently outperforms the evaluated baselines on TruckScenes, demonstrating generalization across different radar configurations and operating ranges.

% A shared representation feeds complementary semantic and query-based instance heads, where semantic foreground evidence directly guides instance segmentation without geometric clustering. An auxiliary RCS-consistency objective further regularizes instance-level reflectivity characteristics.

\section{Related Work}
\label{sec:related_work}

{
\noindent\textit{Radar Representation Learning}:
Automotive radar perception is commonly formulated using either structured bird's-eye-view (BEV) representations or native point-domain processing. BEV methods regularize sparse measurements into grids or pillars and have shown strong performance for 3D object detection. K-Radar~\cite{paek2022k} demonstrated 4D radar detection under diverse weather
conditions, while MMF-BEV~\cite{mayank2026multi} introduced RCS-based radar encoding with safety aware BEV fusion. DRIFT~\cite{pei2026drift} further combines point and pillar representations to capture complementary local and global context. However, rasterization weakens the direct correspondence to individual radar returns required for point-wise semantic and instance prediction.\looseness=-1

Point-based approaches preserve irregular measurement geometry together with radar-specific attributes such as Doppler velocity and RCS. RadarScenes~\cite{schumann2021radarscenes} enabled systematic point-wise semantic evaluation, while GRT~\cite{zeller2023grt} introduced radar-specific attention for sparse semantic segmentation and RVT~\cite{zeller2023rvt} exploited Doppler cues for moving-object segmentation. These methods establish the effectiveness of native radar representations, but predict semantic or motion labels rather than class-aware object instances. Our proposed RPT extends point-based processing to panoptic segmentation, using pose-aligned temporal context to enrich sparse current-scan features.} 

% RPT therefore retains the point domain while incorporating pose-aligned temporal context for joint panoptic reasoning.}

{\parskip=2pt
\noindent\textit{3D Panoptic and Mask-Based Segmentation}:
LiDAR panoptic segmentation provides the closest methodological foundation. Panoptic-PolarNet~\cite{zhou2021panopticpolarnet} and DS-Net~\cite{hong2021dsnet} combine semantic prediction with geometric instance cues and subsequent grouping, while
EfficientLPS~\cite{sirohi2022efficientlps} employs range-aware features and learned panoptic fusion. These approaches benefit from dense LiDAR geometry, where object centers and local spatial structure provide reliable instance cues, an assumption that is considerably weaker for sparse radar returns. Query-based methods reduce the dependency on geometric clustering. Mask3D~\cite{schult2023mask3d} represents instances through learned mask queries, while MaskPLS~\cite{marcuzzi2023maskpls} extends this formulation to LiDAR panoptic segmentation for autonomous driving. Although such set-based prediction is permutation invariant, generic queries are designed for substantially denser point clouds. In radar, severe foreground sparsity and dominant static background make unconstrained object discovery more challenging, motivating semantic guidance during query initialization and refinement.}

{\parskip=2pt
\noindent\textit{Radar Instance and Panoptic Segmentation}: 
Radar instance and panoptic segmentation remain comparatively recent research directions. RIT~\cite{zeller2024rit} exploits temporal measurements for class-agnostic moving-instance segmentation, demonstrating the importance of motion and temporal context. Siddhartha~\emph{et~al.}~\cite{siddhartha2022radarpanoptic} jointly predict semantic labels and instance identifiers, but resolve instance ambiguity through ordered labels rather than permutation-invariant set prediction. SemRaFiner~\cite{zeller2025semrafiner} is the closest radar-specific panoptic approach. It builds on RIT instance predictions and introduces radius-based attention to address spatially varying radar density before semantic-instance refinement. However, instance discovery depends on an upstream class-agnostic instance estimate, and the neighborhood support is not explicitly conditioned on radar reflectivity and viewing geometry. RPT instead jointly couples radar-specific representation learning with semantic-guided set prediction.}

\section{Radar Panoptic Transformer}
\label{sec:method}

\subsection{Problem Formulation and Overview}
\label{sec:overview}

We formulate radar panoptic segmentation directly in the native point domain. At time $t$, the current radar scan is represented as
\begin{equation}\small
\mathcal{X}^{t}
=
\left\{
\mathbf{x}_{i}^{t}
\right\}_{i=1}^{N_t},
\qquad
\mathbf{x}_{i}^{t}
=
\left[
(\mathbf{p}_{i}^{t})^\top,\,
v_{i}^{r,\mathrm{comp},t},\,
r_{i}^{\mathrm{RCS},t}
\right]^\top ,
\label{eq:radar_input}
\end{equation}
where $\mathbf{p}_{i}^{t}=[x_i^t,y_i^t,z_i^t]^\top$ denotes the Cartesian
position of detection $i$ in the current ego frame,
$v_{i}^{r,\mathrm{comp},t}$ its ego-motion-compensated radial Doppler velocity,
$r_{i}^{\mathrm{RCS},t}$ its radar cross section (RCS), and $N_t$ the number
of detections in the current scan. For each detection, RPT predicts a panoptic label $\mathbf{Y}_{i}^{t}=(C_i^t,I_i^t)$ comprising semantic class $C_i^t$ and instance identifier $I_i^t$. We treat static returns as \emph{stuff} with $I_i^t=0$, whereas dynamic \emph{thing} classes receive unique instance identities.

\begin{figure*}[t]
    \centering
    \includegraphics[
        width=\textwidth,
        trim={12 0 0 10},
        clip
    ]{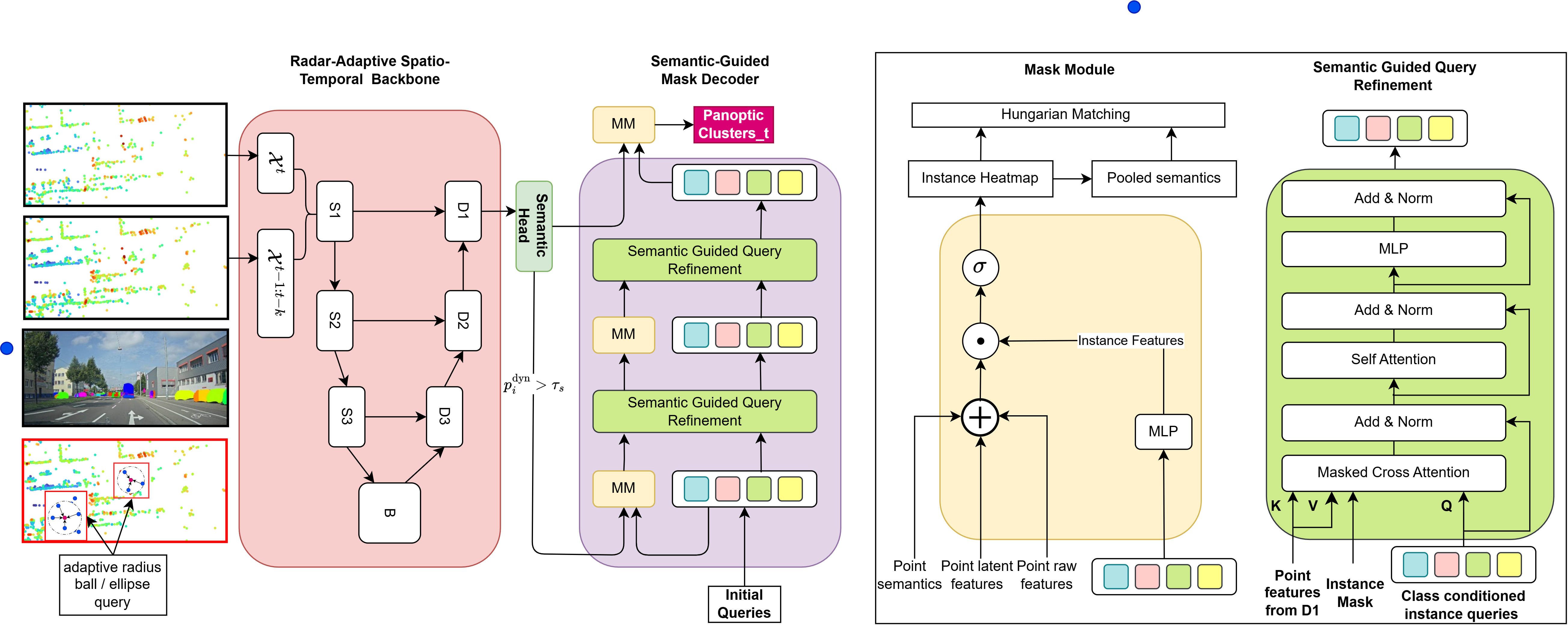}
    \caption{\textbf{Overview of Radar Panoptic Transformer (RPT).}
    The radar-adaptive spatio-temporal backbone first applies shared shallow
    encoding to the current and pose-aligned historical scans, followed by a
    single temporal fusion step into the current-scan representation. The
    resulting features are propagated through the remaining hierarchical
    encoder--decoder. Point-wise semantic predictions provide a dynamic
    foreground prior for semantic-guided query initialization and masked
    refinement. The radar-aware mask module (MM) predicts a class and point-level mask for
    each instance query, which are fused with the semantic branch to obtain the
    final panoptic labels.}
    \label{fig:rpt_architecture}
    \vspace{-0.5cm}
\end{figure*}

As illustrated in Fig.~\ref{fig:rpt_architecture}, RPT comprises three coupled
modules: (i)~a radar-adaptive spatio-temporal backbone implemented as a
hierarchical point-based encoder-decoder, (ii)~a point-wise semantic head, and (iii)~a semantic-guided mask decoder. In the backbone, the current scan defines the prediction domain, while the preceding $T$ scans provide pose-aligned
temporal context. Current and historical detections first pass through the
shared encoder stage S1, after which grouped vector attention fuses historical
features into the current representation. Only the temporally refined
current-scan stream is propagated through the remaining encoder stages and
decoder. The semantic head then produces point-wise class probabilities, which
are additionally used by the mask decoder to initialize and refine instance
queries. The stage-wise backbone configuration is summarized in
Tab.~\ref{tab:architecture}.

\begin{table}[t]
\centering
\footnotesize
\caption{RPT backbone configuration. $N$ denotes the number of current-scan
points. S/D denote encoder/decoder stages. Op., Ch., $K$, and $r_0$ denote
operator, output channels, maximum nearest neighbors, and nominal neighborhood
radius, respectively.}
\label{tab:architecture}

\setlength{\tabcolsep}{2.7pt}
\renewcommand{\arraystretch}{1.03}

\begin{tabular}{cccc}
\toprule
\textbf{Pts.} &
\textbf{Stage} &
\textbf{Encoder: Op./Ch./$K$/$r_0$} &
\textbf{Decoder: Op./Ch./$K$/$r_0$} \\
\midrule

$N$
& S1/D1
& KPConv$\times6$/64/12/5 (shared)
& KPConv$\times1$/64/12/5 \\

$N/2$
& S2/D2
& KPConv$\times2$/128/16/6
& KPConv$\times1$/128/16/6 \\

$N/4$
& S3/D3
& GVA$\times1$/256/24/8
& GVA$\times1$/256/24/8 \\

$N/8$
& S4
& GVA$\times1$/512/24/10
& -- \\

\bottomrule
\end{tabular}

\vspace{-0.35cm}
\end{table}

\subsection{Radar-Adaptive Spatio-Temporal Backbone}
\label{sec:encoder}

{
\noindent\textit{Spatio-Temporal Feature Refinement}:
We first transform the historical radar detections into the current ego frame. Let $\mathbf{T}_{W\leftarrow E_\tau}$ and
$\mathbf{T}_{W\leftarrow E_t}$ denote the ego poses at historical time
$\tau$ and current time $t$, respectively. Using homogeneous coordinates
\begin{equation}
\bar{\mathbf{p}}_{j}^{\,\tau\rightarrow t}
=
\mathbf{T}_{W\leftarrow E_t}^{-1}
\mathbf{T}_{W\leftarrow E_\tau}
\bar{\mathbf{p}}_{j}^{\,\tau},
\label{eq:temporal_alignment}
\end{equation}
where radar-to-ego extrinsics are applied beforehand when required. Only the
Cartesian coordinates are transformed; compensated Doppler and RCS remain
associated measurement attributes. As summarized in Tab.~\ref{tab:architecture}, the current and pose-aligned historical scans are projected into a shared latent space and processed by the
common S1 KPConv stage~\cite{thomas2019kpconv}. Temporal fusion is then
performed once after S1 using grouped vector attention (GVA)~\cite{wu2022point}. For current point $i$ and historical point $j$ from scan $\tau$, the relative displacement is\looseness=-1
\begin{equation}
\Delta\mathbf{p}_{ij}^{\tau}
=
\mathbf{p}_{i}^{t}
-
\mathbf{p}_{j}^{\,\tau\rightarrow t}.
\label{eq:relative_position}
\end{equation}

Following Point Transformer V2~\cite{wu2022point}, the query--key relation is
conditioned on relative position as
\begin{equation}
\mathbf{r}_{ij}^{\tau}
=
\delta_{\mathrm{mul}}(\Delta\mathbf{p}_{ij}^{\tau})
\odot
(\mathbf{q}_i-\mathbf{k}_j)
+
\delta_{\mathrm{bias}}(\Delta\mathbf{p}_{ij}^{\tau}),
\label{eq:gva_relation}
\end{equation}
where $\mathbf{q}_i=W_q\mathbf{h}_i^t$ and
$\mathbf{k}_j=W_k\mathbf{h}_j^\tau$ are learned query and key projections,
$\odot$ denotes the Hadamard product, and
$\delta_{\mathrm{mul}}$ and $\delta_{\mathrm{bias}}$ are MLP-based positional
encoders mapping the relative 3D displacement to the latent relation space.
The multiplicative positional encoding modulates the feature relation
channel-wise, while the additive term provides a geometric bias. The relation is projected to $G_a$ grouped attention weights
\begin{equation}
\boldsymbol{\alpha}_{ij}^{\tau}
=
\operatorname{softmax}_{\mathcal{N}_{i}^{\mathrm{temp}}}
\!\left(
\omega_{G_a}(\mathbf{r}_{ij}^{\tau})
\right),
\label{eq:gva_attention}
\end{equation}
where $\omega_{G_a}$ is a learned grouped weight-encoding function and
$\mathcal{N}_{i}^{\mathrm{temp}}$ denotes the local set of pose-aligned
historical neighbors of current point $i$. The temporally refined current feature is obtained through residual aggregation
\begin{equation}
\widetilde{\mathbf{h}}_{i}^{t}
=
\mathbf{h}_{i}^{t}
+
\sum_{(\tau,j)\in\mathcal{N}_{i}^{\mathrm{temp}}}
\mathcal{B}(\boldsymbol{\alpha}_{ij}^{\tau})
\odot
W_v\mathbf{h}_{j}^{\tau},
\label{eq:temporal_gva}
\end{equation}
where $W_v$ is the value projection and $\mathcal{B}(\cdot)$ broadcasts each
group-wise coefficient over its associated feature channels.} 

{\parskip=2pt
\noindent\textit{Radar-Adaptive Neighborhood Construction}:
The temporally refined feature $\widetilde{\mathbf{h}}_{i}^{t}$ remains
anchored to the current-scan detection $i$ and is propagated through the
subsequent encoder stages. Since automotive radar exhibits strongly non-uniform
sampling with respect to range, viewing geometry, and target reflectivity, a
fixed Euclidean support can be suboptimal across different measurement
conditions. We therefore predict a point-dependent neighborhood radius at
encoder stage $l$ as
\begin{equation}
\hat{r}_{i}^{(l)}
=
r_{\min}^{(l)}
+
\left(r_{\max}^{(l)}-r_{\min}^{(l)}\right)
\sigma\!\left(
\phi_r^{(l)}
[\rho_i,\theta_i,r_i^{\mathrm{RCS}}]
\right),
\label{eq:adaptive_radius}
\end{equation}
where $\rho_i$ and $\theta_i$ denote the range and azimuth of detection $i$
with respect to its originating radar sensor, and
$r_i^{\mathrm{RCS}}$ denotes its reflectivity. The lightweight MLP
$\phi_r^{(l)}$ maps these radar-specific attributes to a scalar radius logit
and is optimized jointly with the backbone. The sigmoid and stage-dependent
bounds $r_{\min}^{(l)}$ and $r_{\max}^{(l)}$ constrain the resulting
ball-query support, allowing sparse long-range detections to aggregate over a
larger spatial neighborhood while retaining localized support for denser
near-range measurements. At most $K$ neighbors are retained for each local
operator, as specified in Tab.~\ref{tab:architecture}.\looseness=-1

{\parskip=2pt
\noindent\textit{Hierarchical Current-Scan Encoding and Decoding}:
After temporal fusion,
$\widetilde{\mathbf{h}}_{i}^{t}$ is propagated from S2 onward. The current-scan
representation is progressively downsampled through the encoder hierarchy
$N\rightarrow N/2\rightarrow N/4\rightarrow N/8$, using the
stage-dependent radar-adaptive neighborhoods described above. The S4
representation at $N/8$ forms the bottleneck. The decoder then restores the
resolution as
$N/8\rightarrow N/4\rightarrow N/2\rightarrow N$
using inverse-distance-weighted interpolation~\cite{qi2017pointnet++} and
encoder-decoder skip connections.}

{\parskip=2pt
\noindent\textit{Backbone Output and Semantic Prediction}:
The decoder yields the full-resolution current-scan representation as
\begin{equation}
\mathbf{F}
=
[\mathbf{f}_1,\ldots,\mathbf{f}_N]^\top
\in\mathbb{R}^{N\times D},
\label{eq:backbone_output}
\end{equation}
where $\mathbf{f}_i$ denotes the final feature associated with current
detection $i$. A point-wise semantic head maps these features to
$\mathbf{p}_i^s=\operatorname{softmax}(\mathbf{s}_i)$, providing both the
semantic classes and the dynamic foreground prior used by the
semantic-guided mask decoder described next.}

\subsection{Semantic-Guided Mask Decoder}
\label{sec:semantic_query}

Following the mask-classification paradigm of
Mask2Former~\cite{cheng2022masked} and
Mask3D~\cite{schult2023mask3d}, RPT represents each candidate dynamic instance
by an object query. In contrast to generic query decoders, semantic information
is introduced explicitly at three stages: query initialization, mask formation,
and iterative query refinement. This coupling is designed for radar point
clouds, where dynamic objects may contain only a few detections within a
strongly dominant static background. The semantic head maps the full-resolution backbone feature $\mathbf{f}_i$ to
point-wise class probabilities
$\mathbf{p}_i^s=\operatorname{softmax}(\mathbf{s}_i)$. We define the dynamic
foreground probability as
\begin{equation}
p_i^{\mathrm{dyn}}
=
1-p_i^s(c_{\mathrm{static}})
=
\sum_{c\in\mathcal{C}_{\mathrm{thing}}}p_i^s(c).
\label{eq:dynamic_probability}
\end{equation}
Detections satisfying $p_i^{\mathrm{dyn}}>\tau_s$ form the candidate set for
query initialization. Importantly, this threshold is used only to select query
seeds. All current-scan points remain available to the subsequent mask
prediction and cross-attention operations. Consequently, weakly classified
radar returns can still be incorporated into an instance during refinement.
For a spatially distributed seed $r_j$, the initial query is constructed as
\begin{equation}
\mathbf{q}_j^0
=
\mathbf{q}_j^{\mathrm{learn}}
+
\phi_{\mathrm{seed}}
\!\left(
\mathbf{f}_{r_j}
\Vert
\mathbf{p}_{r_j}^{s}
\Vert
\mathbf{x}_{r_j}^{t}
\right),
\quad j=1,\ldots,Q,
\label{eq:semantic_query_init}
\end{equation}
where $\mathbf{q}_j^{\mathrm{learn}}$ is a learned query prior,
$\Vert$ denotes concatenation, and $\phi_{\mathrm{seed}}$ projects the
backbone, semantic, and raw radar features to the query dimension $d_q$.
The initialization therefore biases the finite query set toward measurements
with dynamic semantic evidence while retaining object-level learned priors.

\paragraph{Radar-Aware Mask Module}
At every decoder stage, the mask module (MM) converts the current object
queries into point-level instance hypotheses. As illustrated in
Fig.~\ref{fig:rpt_architecture}, the module combines three complementary
representations of radar detection $i$: the backbone feature $\mathbf{f}_i$,
the semantic probability vector $\mathbf{p}_i^s$, and the raw radar
measurement $\mathbf{x}_i^t$. Since these quantities have different
dimensions, they are independently projected to a common mask-feature space
and combined as
\begin{equation}
\mathbf{u}_i
=
\phi_f(\mathbf{f}_i)
+
\phi_s(\mathbf{p}_i^s)
+
\phi_x(\mathbf{x}_i^t),
\label{eq:mm_point_feature}
\end{equation}
where $\phi_f$, $\phi_s$, and $\phi_x$ are learned projections. Thus,
$\mathbf{u}_i$ contains learned geometric and temporal context from the
backbone together with explicit semantic and radar-measurement cues. For query $\mathbf{q}_j^{l-1}$, an MLP generates the corresponding
instance embedding
$\mathbf{e}_j^{l-1}=\phi_q(\mathbf{q}_j^{l-1})$.
The query-specific instance heatmap is obtained from point--query similarity:
\begin{equation}
\widehat{m}_{ij}^{\,l-1}
=
\sigma
\!\left(
\mathbf{u}_i^\top\mathbf{e}_j^{l-1}
\right),
\label{eq:mask_probability}
\end{equation}
where $\sigma(\cdot)$ denotes the sigmoid function. Each query therefore
induces a soft mask over the current radar scan rather than relying on
distance-based clustering. The semantic evidence associated with an instance is obtained directly from
the point-wise semantic branch by mask-weighted pooling:
\begin{equation}
\bar{\mathbf{p}}_{j}^{\,s,l-1}
=
\frac{
\sum_{i=1}^{N}
\widehat{m}_{ij}^{\,l-1}\mathbf{p}_i^s
}{
\sum_{i=1}^{N}
\widehat{m}_{ij}^{\,l-1}+\epsilon
}.
\label{eq:pooled_semantics}
\end{equation}
This operation aggregates semantic evidence only from points supported by the
predicted instance mask. Compared with predicting the instance class solely
from a generic query embedding, the class estimate is therefore explicitly
grounded in the semantic predictions of the radar detections assigned to that
instance. The pooled semantic representation is subsequently used to condition the
query before the next refinement stage:
\begin{equation}
\widetilde{\mathbf{q}}_j^{\,l-1}
=
\mathbf{q}_j^{l-1}
+
\phi_c
\!\left(
\bar{\mathbf{p}}_{j}^{\,s,l-1}
\right),
\label{eq:class_conditioned_query}
\end{equation}
where $\phi_c$ projects the pooled semantic evidence to the query dimension.
The resulting $\widetilde{\mathbf{q}}_j^{\,l-1}$ constitutes the
class-conditioned instance query shown in
Fig.~\ref{fig:rpt_architecture}.

\paragraph{Semantic-Guided Query Refinement}
The predicted instance mask restricts the support of the subsequent
cross-attention operation. We use the soft mask as an additive attention bias,
\begin{equation}
B_{ji}^{l}
=
\log
\!\left(
\max(\widehat{m}_{ij}^{\,l-1},\epsilon_m)
\right),
\label{eq:mask_attention_bias}
\end{equation}
which favors points already associated with the query while retaining access
to weakly supported detections.

As shown in Fig.~\ref{fig:rpt_architecture}, the class-conditioned query forms
the attention query, while the full-resolution backbone features provide the
keys and values. The cross-attention logit is
\begin{equation}
A_{ji}^{l}
=
\frac{
(W_Q\widetilde{\mathbf{q}}_j^{\,l-1})^\top
(W_K\mathbf{f}_i)
}{
\sqrt{d_a}
}
+
B_{ji}^{l},
\label{eq:semantic_mask_attention}
\end{equation}
where $W_Q$ and $W_K$ are learned projections and $d_a$ is the attention
dimension. Semantic guidance therefore enters the refinement through the
class-conditioned query, while the predicted instance mask supplies its
spatial support.

After normalization over the current-scan points, the attention weights
aggregate the corresponding value features $W_V\mathbf{f}_i$. The resulting
query is processed by residual normalization, query self-attention, and an
MLP. The MM and semantic-guided refinement blocks are repeated over $L$
decoder layers, such that each stage updates the instance mask, pools its
semantic evidence and refines the corresponding object query. The final
decoder output consists of the instance masks
$\widehat{\mathbf{M}}_j\in[0,1]^N$ and their associated instance-level
semantic predictions.\looseness=-1

\subsection{Training Objective and Panoptic Fusion}
\label{sec:training}

Training jointly optimizes point-wise semantic segmentation and
permutation-invariant instance prediction. The unordered set of ground-truth
dynamic instances is
\begin{equation}
\mathcal{Y}
=
\left\{
(\mathbf{M}_g,c_g)
\right\}_{g=1}^{N_{\mathrm{gt}}},
\label{eq:gt_instances}
\end{equation}
where $\mathbf{M}_g\in\{0,1\}^{N}$ denotes the binary point mask of instance
$g$, $c_g\in\mathcal{C}_{\mathrm{thing}}$ its semantic class, and
$N_{\mathrm{gt}}$ the number of annotated dynamic instances. For each predicted query, the final mask module provides an instance mask
$\widehat{\mathbf{M}}_j$ together with instance-level semantic evidence
$\mathbf{p}_j^q$ obtained from the mask-pooled point semantics in
Eq.~\eqref{eq:pooled_semantics}. During training, the $Q$ predictions are
matched one-to-one with the ground-truth instances using Hungarian matching,
following DETR~\cite{carion2020end} and
Mask3D~\cite{schult2023mask3d}. The pairwise matching cost is
\begin{equation}
\mathcal{C}_{gj}^{\mathrm{match}}
=
-\alpha_{\mathrm{cls}}\log p_j^q(c_g)
+\alpha_{\mathrm{bce}}\mathcal{C}_{gj}^{\mathrm{bce}}
+\alpha_{\mathrm{dice}}\mathcal{C}_{gj}^{\mathrm{dice}}.
\label{eq:matching_cost}
\end{equation}
The classification term measures agreement between the mask-pooled semantic
prediction and the ground-truth class, while the BCE and Dice terms compare
$\widehat{\mathbf{M}}_j$ with $\mathbf{M}_g$. Dice supervision is particularly
useful for sparse radar targets represented by only one or a few detections. The semantic branch is 
optimized using focal-Tversky~\cite{abraham2019novel} and Lov\'asz~\cite{berman2018lovasz} losses
\begin{equation}
\mathcal{L}_{\mathrm{sem}}
=
\lambda_{\mathrm{FT}}\mathcal{L}_{\mathrm{FT}}
+
\lambda_{\mathrm{Lov}}\mathcal{L}_{\mathrm{Lov}},
\label{eq:semantic_loss}
\end{equation}
to address foreground--background imbalance and improve class-wise overlap.

\paragraph{RCS-Consistent Instance Supervision}
Spatial mask supervision does not explicitly constrain instance-level radar
reflectivity. For each predicted mask $\widehat{\mathbf{M}}_j$, we compute the
mask-weighted RCS mean $\widehat{\mu}_j^{\mathrm{RCS}}$ and variance
$\widehat{\sigma}_j^{2,\mathrm{RCS}}$, with corresponding ground-truth
statistics $\mu_g^{\mathrm{RCS}}$ and $\sigma_g^{2,\mathrm{RCS}}$. Given the
Hungarian assignment $\widehat{\pi}(g)$, we define
\begin{equation}\scriptsize
\mathcal{L}_{\mathrm{RCS}}
=
\frac{1}{N_{\mathrm{gt}}}
\sum_{g=1}^{N_{\mathrm{gt}}}
\left(
\left|
\widehat{\mu}_{\widehat{\pi}(g)}^{\mathrm{RCS}}
-\mu_g^{\mathrm{RCS}}
\right|
+
\lambda_{\mathrm{var}}
\left|
\widehat{\sigma}_{\widehat{\pi}(g)}^{2,\mathrm{RCS}}
-\sigma_g^{2,\mathrm{RCS}}
\right|
\right).
\label{eq:rcs_loss}
\end{equation}
This radar-native regularizer encourages matched instances to preserve their
RCS mean and variance, complementing other losses. The complete objective is:
\begin{equation}\scriptsize
\mathcal{L}
=
\lambda_{\mathrm{sem}}\mathcal{L}_{\mathrm{sem}}
+\lambda_{\mathrm{cls}}\mathcal{L}_{\mathrm{cls}}
+\lambda_{\mathrm{bce}}\mathcal{L}_{\mathrm{bce}}
+\lambda_{\mathrm{dice}}\mathcal{L}_{\mathrm{dice}}
+\lambda_{\mathrm{RCS}}\mathcal{L}_{\mathrm{RCS}}.
\label{eq:total_loss}
\end{equation}

\paragraph{Panoptic Fusion}
Hungarian matching is required only during training. At inference, low-confidence
instance hypotheses are discarded. For each retained query $j$, let
$\widehat{c}_j=\arg\max_{c\in\mathcal{C}_{\mathrm{thing}}}p_j^q(c)$ denote
the class obtained from its mask-pooled semantic evidence. The point-wise
\emph{thing} score is
\begin{equation}
S_{ij}^{\mathrm{thing}}
=
\widehat{m}_{ij}\,
p_j^q(\widehat{c}_j)\,
p_i^s(\widehat{c}_j),
\label{eq:thing_score}
\end{equation}
combining instance-mask confidence, instance-level semantic confidence, and
the local semantic evidence of detection $i$. A point is assigned to its
highest-scoring dynamic query when the corresponding thing score exceeds its
static semantic score $p_i^s(c_{\mathrm{static}})$. Otherwise, it retains the
static label with instance ID zero. RPT therefore produces class-aware
point-level panoptic predictions directly in the radar domain without
geometric clustering or bounding-box post-processing.

\begin{table}[t]
\centering
\caption{Comparison of automotive datasets.}
\label{tab:dataset_comparison}
\setlength{\tabcolsep}{2.5pt}
\renewcommand{\arraystretch}{1.08}
\resizebox{\columnwidth}{!}{%
\begin{tabular}{lcccc}
\toprule
\textbf{Dataset}
& \textbf{Radar setup}
& \textbf{Coverage}
& \textbf{Range}
& \textbf{Supervision} \\
\midrule

RadarScenes~\cite{schumann2021radarscenes}
& 4$\times$ radar
& front
& 100\,m
& point cls. + track ID \\

NuScenes~\cite{fong2022panoptic}
& 5$\times$ radar
& $360^{\circ}$
& $50$\,m
& 3D boxes + track ID \\

VoD~\cite{palffy2022multi}
& 1$\times$ 3+1D radar
& front
& $50$\,m
& 3D boxes + track ID \\

TruckScenes~\cite{fent2024mantruckscenes}
& 6$\times$ 4D radar
& $360^{\circ}$
& $230$\,m
& 3D boxes + track ID \\

\bottomrule
\end{tabular}%
}
\vspace{-0.4cm}
\end{table}

\begin{table*}[t]
    \centering
    \caption{Panoptic segmentation results on RadarScenes Test set. RadarScenes baseline results are reproduced from SemRaFiner~\cite{zeller2025semrafiner}. Gray rows denote methods using temporal/multi-scan input. All values are percentages.}
    \label{tab:radar_panoptic_results}
    \setlength{\tabcolsep}{3.0pt}
    \renewcommand{\arraystretch}{1.10}

    \resizebox{\textwidth}{!}{%
    \begin{tabular}{llcc*{12}{c}}
        \toprule
        \multirow{2}{*}{Method}
        & \multirow{2}{*}{Input}
        & \multirow{2}{*}{PQ}
        & \multirow{2}{*}{mIoU}
        & \multicolumn{6}{c}{PQ}
        & \multicolumn{6}{c}{IoU} \\
        \cmidrule(lr){5-10}
        \cmidrule(lr){11-16}
        & & & &
        static & car & ped. & ped.\ grp. & bike & truck &
        static & car & ped. & ped.\ grp. & bike & truck \\
        \midrule

        \rowcolor{gray!20}
        RadarPNv1~\cite{schumann2018semanticradar}
        & \multirow{3}{*}{multi-scan}
        & -- & 61.0
        & -- & -- & -- & -- & -- & --
        & 98.7 & 58.2 & 36.0 & 58.7 & 58.4 & 56.1 \\

        \rowcolor{gray!20}
        RadarPNv2~\cite{schumann2019scene}
        &
        & -- & 61.9
        & -- & -- & -- & -- & -- & --
        & 98.7 & 63.8 & 38.8 & 58.5 & 51.0 & 61.0 \\

        \rowcolor{gray!20}
        STA-Net~\cite{zhang2023spatial}
        &
        & -- & 70.4
        & -- & -- & -- & -- & -- & --
        & 99.8 & 75.9 & 43.1 & 82.0 & 67.1 & 54.6 \\

        \midrule

        Mask3D~\cite{schult2023mask3d}
        & \multirow{3}{*}{single-scan}
        & 56.9 & 56.1
        & 98.4 & 70.6 & 18.4 & 31.9 & 57.7 & 64.3
        & 98.6 & 68.7 & 11.8 & 35.9 & 57.7 & 63.9 \\

        GRT~\cite{zeller2023grt} + DBSCAN~[7]
        &
        & 56.3 & 56.9
        & 98.7 & 59.0 & 29.7 & 38.3 & 54.6 & 57.3
        & 98.7 & 58.5 & 21.7 & 46.0 & 54.6 & 61.8 \\

        SemRaFiner~\cite{zeller2025semrafiner}
        &
        & 81.4 & 70.4
        & \textbf{99.7} & 85.7 & \textbf{59.2}
        & 83.1 & 78.9 & 82.1
        & 99.4 & 74.9 & \textbf{42.1}
        & 71.0 & 65.2 & \textbf{69.6} \\

        \textbf{RPT (ours)}
        & 
        & \textbf{82.2} & \textbf{71.6}
        & 99.5 & \textbf{87.9} & 57.4
        & \textbf{85.8} & \textbf{80.1} & \textbf{82.5}
        & \textbf{99.5} & \textbf{76.0} & 40.5
        & \textbf{74.8} & \textbf{68.2} & 68.6 \\

        \bottomrule
    \end{tabular}%
    }
    \vspace{-0.3cm}
\end{table*}

\begin{table}[t]
    \centering
    \caption{Panoptic segmentation results on the MAN TruckScenes validation
    set.}
    \label{tab:truckscenes_panoptic_results}

    \setlength{\tabcolsep}{4.0pt}
    \renewcommand{\arraystretch}{1.10}

    \resizebox{\columnwidth}{!}{%
    \begin{tabular}{llccccc ccc}
        \toprule
        \multirow{2}{*}{Method}
        & \multirow{2}{*}{Input}
        & \multirow{2}{*}{PQ}
        & \multirow{2}{*}{mIoU}
        & \multicolumn{3}{c}{PQ}
        & \multicolumn{3}{c}{IoU} \\
        \cmidrule(lr){5-7}
        \cmidrule(lr){8-10}
        & & & &
        static & car & truck &
        static & car & truck \\
        \midrule

        \rowcolor{gray!20}
        STA-Net~\cite{zhang2023spatial}
        & multi-scan
        & -- & 72.9
        & -- & -- & --
        & 96.8 & 58.1 & 63.9 \\

        PTv3~\cite{wu2024ptv3}
        & 
        & 54.5 & 66.4
        & 98.1 & 39.6 & 25.9
        & 94.7 & 50.2 & 54.3 \\

        Mask3D~\cite{schult2023mask3d}
        & single-scan
        & 56.3 & 68.8
        & 98.4 & 42.8 & 27.6
        & 95.2 & 53.5 & 57.7 \\

        \textbf{RPT (ours)}
        & 
        & \textbf{62.8} & \textbf{73.3}
        & \textbf{98.4} & \textbf{50.6} & \textbf{39.4}
        & \textbf{96.9} & \textbf{58.5} & \textbf{64.6} \\

        \bottomrule
    \end{tabular}%
    }
    \vspace{-0.3cm}
\end{table}

\section{Experimental Evaluation}
\label{sec:experiments}

We evaluate RPT on two automotive radar datasets and assess its semantic and panoptic segmentation performance. We first describe the datasets, their class distributions, and evaluation metrics in Sec.~\ref{sec:datasets},
followed by the implementation and training setup in
Sec.~\ref{sec:experimental_setup}. We then present benchmarking results in
Sec.~\ref{sec:benchmarking}, analyze the contributions of individual model
components through ablation studies in Sec.~\ref{sec:ablations}, runtime analysis in Sec.~\ref{sec:runtime}, and conclude
with qualitative analysis in Sec.~\ref{sec:qualitative_evaluation}.

\subsection{Datasets and Evaluation}
\label{sec:datasets}

RadarScenes offers point-wise semantic labels and track identities,
enabling direct panoptic supervision. TruckScenes provides six surround 4D
radars with nearly $360^{\circ}$ coverage and ranges up to
$230\,\mathrm{m}$. Since only 3D boxes are annotated, radar detections are
temporally aligned to the annotation timestamp and labeled by point-in-box
association. In comparison, nuScenes~\cite{fong2022panoptic} provides
comparatively sparse, filtered radar detections, while
VoD~\cite{palffy2022multi} is limited to a single forward-facing 4D radar.
Our selected datasets are strongly background dominated, with static returns accounting
for $96.7\%$ of RadarScenes and $95.8\%$ of TruckScenes. RadarScenes dynamic
classes comprise approximately $0.17$--$1.42\%$ of total radar detections and include vehicles and VRUs. $17.31\%$ of the instances are singletons. TruckScenes contains $1.37\%$ car, $2.76\%$ truck, $0.02\%$ pedestrian, and $0.01\%$
two-wheeler returns. Since distant objects are often represented by only a few
detections, and pedestrian/two-wheeler samples are extremely limited in
TruckScenes, we restrict evaluation to cars and trucks. We report mIoU for semantic segmentation and PQ for panoptic segmentation~\cite{vodisch2023codeps}.

\subsection{Experimental Setup}
\label{sec:experimental_setup}

For RadarScenes, we follow the split of~\cite{zeller2024rit}
with 130/6/22 train/validation/test sequences and cap each scan at $N=1024$
points (569 on average). For TruckScenes, we use $N=4096$ for an average of
2841 points per scan. We use a common query budget of $Q=64$ for both datasets. We project radar features to a $D=32$-dimensional latent space. Spatio-temporal attention uses $G_a=4$ groups, while the semantic-guided decoder employs $L=2$ refinement layers with a seed threshold of $\tau_s=0.40$. We provide temporal context with $T=2$ pose-compensated historical scans. We train with a batch size of 128 and Focal-Tversky parameters
$\alpha=0.3$, $\beta=0.7$, and $\gamma=2.0$, with
$\lambda_{\mathrm{FT}}=\lambda_{\mathrm{Lov}}=0.5$. Training spans 80 epochs: the semantic backbone and head are optimized for
40 epochs, after which they are frozen and the mask decoder is trained for the remaining 40 epochs.
We use an initial learning rate of $5\times10^{-4}$ with cosine annealing.
The loss weights are
$\lambda_{\mathrm{sem}}=0.25$,
$\lambda_{\mathrm{cls}}=1.0$,
$\lambda_{\mathrm{bce}}=4.0$,
$\lambda_{\mathrm{dice}}=3.0$, and
$\lambda_{\mathrm{RCS}}=2.0$.

\subsection{Benchmarking Results}
\label{sec:benchmarking}

\begin{figure}[htbp]
    \centering
    \includegraphics[
        width=\columnwidth,
        trim={15 15 0 6.2},
        clip
    ]{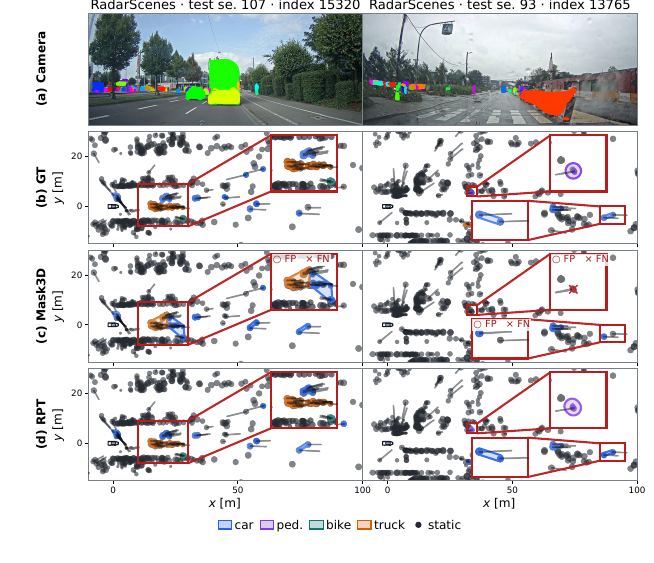}
    \caption{\textbf{Qualitative comparison on RadarScenes test set.}
    Left: Mask3D merges/fragmentarily assigns adjacent-lane vehicles
    with similar radial motion, whereas RPT preserves two distinct and coherent
    instances. Right: In a rainy construction area scene, Mask3D misses a
    pedestrian and misclassifies a sparsely observed moving car as static. RPT correctly recovers both the pedestrian and car instances. Rows show
    (a) camera view, (b) ground truth, (c) Mask3D, and (d) RPT predictions.}
    \label{fig:radarscenes_qualitative}
    \vspace{-0.4cm}
\end{figure}

\begin{figure}[t]
    \centering
    \includegraphics[
        width=\columnwidth,
        trim={5 20 0 15},
        clip
    ]{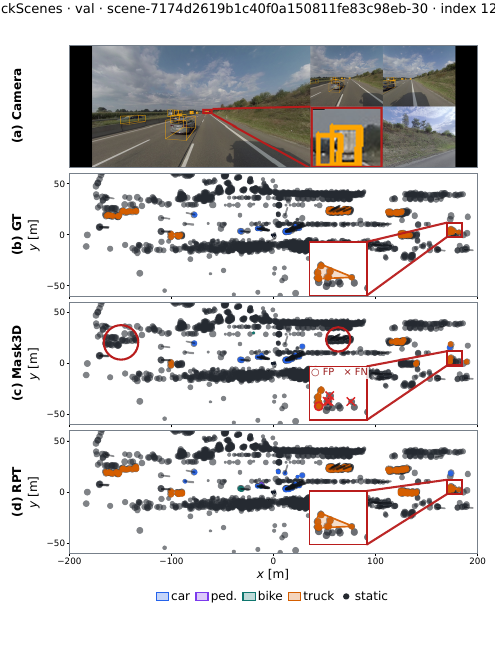}
    \caption{\textbf{Qualitative comparison on a TruckScenes highway scene.}
    (a)~Multi-camera view with an enlarged region of interest.
    (b)~Ground-truth panoptic labels.
    (c)~Mask3D shows FP/FN errors and fragmented long-range instances, including missed trucks circled in red.
    (d)~RPT produces coherent truck instances at longitudinal ranges of up to approximately $180\,\mathrm{m}$ in the front and rear, closely matching the ground truth.}
    \label{fig:truckscenes_qualitative}
    \vspace{-0.3cm}
\end{figure}

\begin{table}[t]
    \centering
    \caption{Ablation of encoder components.}
    \label{tab:architecture_component_ablation}

    \setlength{\tabcolsep}{3pt}
    \renewcommand{\arraystretch}{1.10}

    \begin{tabular}{lc}
        \toprule
        Configuration & mIoU \\
        \midrule
        No temporal module ($T=0$)                 & 67.7 \\
        Temporal + fixed-radius query         & 70.9 \\
        Temporal + learned-radius query       & \textbf{71.6} \\
        \bottomrule
    \end{tabular}
    \vspace{-0.6cm}
\end{table}

\begin{table}[t]
    \centering
    \caption{Ablation of instance decoder.}
    \label{tab:panoptic_component_ablation}
    \setlength{\tabcolsep}{3pt}
    \renewcommand{\arraystretch}{1.10}
    \begin{tabular}{lc}
        \toprule
        Configuration & PQ \\
        \midrule

        Semantic Head + MeanShift Clustering~\cite{cheng1995mean}
        & 73.2 \\

        Semantic Head + semantic-guided decoder
        & 81.6 \\

        Semantic Head + semantic-guided decoder + RCS loss
        & \textbf{82.2} \\
        
        \bottomrule
    \end{tabular}
    \vspace{-0.6cm}
\end{table}

\begin{table}[t]
    \centering
    \caption{Effect of the number of historical radar scans $T$.}
    \label{tab:temporal_ablation}
    \setlength{\tabcolsep}{8pt}
    \renewcommand{\arraystretch}{1.08}
    \begin{tabular}{lcc}
        \toprule
        Temporal input & mIoU & PQ \\
        \midrule
        $T=1$ & 69.4 & 80.0 \\
        $T=2$ (RPT) & \textbf{71.6} & \textbf{82.2} \\
        $T=3$ & 71.5 & 81.8 \\
        \bottomrule
    \end{tabular}
    \vspace{-0.6cm}
\end{table}

\begin{table}[t]
    \centering
    \caption{Model inference runtime on RadarScenes. The baselines of other architectures are taken from Zeller~et~al.~\cite{zeller2025semrafiner}}
    \label{tab:runtime}
    \setlength{\tabcolsep}{5pt}
    \renewcommand{\arraystretch}{1.08}
    \begin{tabular}{lccc}
        \toprule
        \textbf{Method} & \textbf{Params. [M]} & \textbf{Runtime [ms]} & \textbf{PQ}\\
        \midrule
        STA-Net~\cite{zhang2023spatial}
        & 7.4 & 34.9 & -- \\

        RadarPNv1~\cite{schumann2018semanticradar}
        & 1.8 & 278.1 & -- \\

        Mask3D~\cite{schult2023mask3d}
        & 39.6 & 85.2 & 56.9 \\

        GRT~\cite{zeller2023grt}
        & 8.4 & \textbf{24.0} & 56.3\\

        RIT~\cite{zeller2024rit}
        & 3.8 & 31.7 & -- \\

        SemRafiner~\cite{zeller2025semrafiner}
        & 4.5 & 42.1 & 81.4 \\

        \textbf{RPT (ours)}
        & \textbf{5.4} & 32.1 & \textbf{82.2} \\

        \bottomrule
    \end{tabular}
    \vspace{-0.6cm}
\end{table}

Tab.~\ref{tab:radar_panoptic_results} reports the quantitative results. On RadarScenes, RPT achieves $71.6\%$ mIoU and $82.2\%$ PQ, outperforming SemRaFiner~\cite{zeller2025semrafiner} by $1.2$ and $0.8$ percentage points, respectively, and thus achieving a new state of the art. The largest instance-level improvements are obtained for cars, pedestrian groups, bikes, and trucks, indicating that semantic-guided mask prediction is particularly effective for sparse dynamic targets. RPT also outperforms methods that directly aggregate multiple radar scans, including RadarPNv1~\cite{schumann2018semanticradar}, RadarPNv2~\cite{schumann2019scene}, and STA Net~\cite{zhang2023spatial}. 

We further evaluate RPT on MAN TruckScenes. For semantic
segmentation, we select STA-Net as the strongest radar semantic baseline
considered in our evaluation. For panoptic segmentation, we use Mask3D~\cite{schult2023mask3d} and PTv3~\cite{wu2024ptv3} as the principal comparisons because their publicly available implementations enable consistent training and evaluation on the TruckScenes radar data. TruckScenes evaluation further demonstrates RPT's robustness beyond the RadarScenes operating regime and under a substantially different surround-radar configuration. As shown in Tab.~\ref{tab:truckscenes_panoptic_results}, RPT achieves the best overall performance with $73.3\%$ mIoU and $62.8\%$ PQ, outperforming STA-Net in
semantic segmentation and exceeding Mask3D by $6.5$ PQ points. The improvement is particularly pronounced for dynamic vehicles, with car and truck PQ gains of $7.8$ and $11.8$ points over Mask3D, respectively.\looseness=-1

\subsection{Ablation Studies}
\label{sec:ablations}

We analyze the contributions of the radar-adaptive encoder, semantic-guided
instance decoder, and temporal context on RadarScenes.

\subsubsection{Radar-Adaptive Encoder}
Tab.~\ref{tab:architecture_component_ablation} evaluates the feature extraction
and neighborhood construction used by the encoder. Without temporal feature
fusion, RPT obtains $67.7\%$ mIoU. Introducing pose-aligned temporal encoding
with fixed-radius neighborhoods increases performance to $70.9\%$ mIoU.
Replacing the fixed radius neighborhood with the proposed learned neighborhood further improves mIoU to $71.6\%$.
These results show that temporal feature enrichment provides the dominant gain,
while adapting the receptive field to radar-dependent sampling provides an
additional improvement.

\subsubsection{Instance Refinement}
Tab.~\ref{tab:panoptic_component_ablation} isolates the contribution of the
instance prediction stage. Replacing the proposed semantic-guided query decoder
with MeanShift clustering~\cite{cheng1995mean} yields $73.2\%$ PQ, whereas
semantic-guided mask prediction increases PQ to $81.6\%$. The $8.4$-point gain
shows the benefit of integrating semantic evidence directly into instance
formation rather than applying geometric clustering to semantic predictions.
Adding the RCS-consistency objective further improves PQ by $0.6$ points to
$82.2\%$, indicating complementary benefit from radar-native instance-level
reflectivity supervision.

\subsubsection{Temporal Context}
Tab.~\ref{tab:temporal_ablation} studies the temporal horizon while retaining
the complete RPT architecture. Increasing the number of pose-aligned historical
scans from $T=1$ to $T=2$ improves mIoU from $69.4\%$ to $71.6\%$ and PQ from
$80.0\%$ to $82.2\%$. Extending the context to $T=3$ provides no further
improvement, suggesting that additional historical measurements contribute
limited information.

\subsection{Runtime}
\label{sec:runtime}
Tab.~\ref{tab:runtime} compares model complexity and inference latency.
RPT contains $5.4$\,M parameters and processes each scan in $32.1$\,ms on an Nvidia
RTX A6000, corresponding to approximately $31$\,Hz and $24\%$ lower latency
than SemRaFiner. Despite temporal feature extraction and mask-based panoptic
prediction, RPT remains competitive with lighter semantic-only baselines.
Its inference rate also exceeds the $17$\,Hz and $20$\,Hz radar frame rates of
the evaluated datasets, supporting real-time operation while maintaining
benchmark performance.

\subsection{Qualitative Evaluation}
\label{sec:qualitative_evaluation}
Fig.~\ref{fig:radarscenes_qualitative} and
Fig.~\ref{fig:truckscenes_qualitative} highlight the radar-specific advantages of
RPT over the generic Mask3D formulation. Mask3D relies on voxelized sparse
features and generic learnable queries, without explicitly modeling radar
temporal continuity, range-dependent sampling, or reflectivity. In dense traffic cases as in RadarScenes, these mechanisms help separate adjacent vehicles with similar Doppler
signatures and recover sparsely observed dynamic targets that Mask3D misses or
assigns to static background. Semantic-guided query initialization and
semantic-compatible masked attention further focus instance discovery on sparse
foreground returns. On TruckScenes, the same radar-adaptive temporal encoding
enables coherent truck masks in the front and rear longitudinal distance of approximately $180\,\mathrm{m}$ while Mask3D exhibits incomplete masks and false foreground assignments.

\section{Conclusion}
\label{sec:conclusion}

We presented \emph{Radar Panoptic Transformer (RPT)}, a lightweight and
computationally efficient framework for joint semantic and instance
understanding directly from automotive radar point clouds. RPT combines pose-aligned temporal context, radar-adaptive neighborhoods conditioned on range, azimuth, and RCS, enhanced with semantic-guided mask queries for instance formation without explicit geometric clustering. The model achieves state-of-the-art radar panoptic segmentation on RadarScenes and, to the best of our knowledge, provides the first reported radar panoptic evaluation on the long-range, full-surround MAN TruckScenes setting, where it consistently outperforms the evaluated baselines. These results show that RPT provides an effective and practical representation for radar-based dynamic scene understanding across diverse autonomous-driving scenarios.\looseness=-1


%%%%%%%%%%%%%%%%%%%%%%%%%%%%%%%%%%%%%%%%%%%%%%%%%%%%%%%%%%%%%%%%%%%%%%%%%%%%%%%%

%%%%%%%%%%%%%%%%%%%%%%%%%%%%%%%%%%%%%%%%%%%%%%%%%%%%%%%%%%%%%%%%%%%%%%%%%%%%%%%%



% \section*{ACKNOWLEDGMENT}

% The preferred spelling of the word ÒacknowledgmentÓ in America is without an ÒeÓ after the ÒgÓ. Avoid the stilted expression, ÒOne of us (R. B. G.) thanks . . .Ó  Instead, try ÒR. B. G. thanksÓ. Put sponsor acknowledgments in the unnumbered footnote on the first page.

%%%%%%%%%%%%%%%%%%%%%%%%%%%%%%%%%%%%%%%%%%%%%%%%%%%%%%%%%%%%%%%%%%%%%%%%%%%%%%%%

\bibliographystyle{IEEEtran}
\bibliography{references}

\end{document}